\documentclass[times,twocolumn,final,authoryear]{elsarticle}

\usepackage{prletters}
\usepackage{framed,multirow}

\usepackage{amssymb}
\usepackage{latexsym}

\usepackage{url}
\usepackage{xcolor}
\definecolor{newcolor}{rgb}{.8,.349,.1}

\usepackage[hidelinks]{hyperref}
\usepackage[utf8]{inputenc}
\usepackage[small]{caption}
\usepackage{graphicx}
\usepackage{amsmath}
\usepackage{amsthm}
\usepackage{booktabs}
\usepackage{algorithm}
\usepackage{algorithmic}
\usepackage{amsfonts}
\usepackage{multirow}
\journal{Pattern Recognition Letters}

\begin{document}



\begin{frontmatter}

\title{MOON: Multi-Hash Codes Joint Learning for Cross-Media Retrieval}

\author[1]{Donglin \snm{Zhang}}
\author[1]{Xiao-Jun \snm{Wu}\corref{cor1}}
\cortext[cor1]{Corresponding author:}
\ead{wu\_xiaojun@jiangnan.edu.cn}
\author[1]{He-Feng \snm{Yin}}
\author[2]{Josef \snm{Kittler}}

\address[1]{School of Artificial Intelligence and Computer Science , Jiangnan University, 214122, Wuxi, China}
\address[2]{Center for Vision, Speech and Signal Processing (CVSSP), University of Surrey, UK}

\received{}
\finalform{}
\accepted{}
\availableonline{}
\communicated{}

\begin{abstract}
In recent years, cross-media hashing technique has attracted increasing attention for its high computation efficiency and low storage cost.
However, the existing approaches still have some limitations, which need to be explored.
1) A fixed hash length (\emph{e.g.,} 16bits or 32bits) is predefined before learning the binary codes.
Therefore, these models need to be retrained when the hash length changes, that consumes additional computation power, reducing the scalability in practical applications.
2) Existing cross-modal approaches only explore the information in the original multimedia data to perform the hash learning, without exploiting the semantic information contained in the learned hash codes.
To this end, we develop a novel \textbf{M}ultiple hash c\textbf{O}des j\textbf{O}int lear\textbf{N}ing method (MOON) for cross-media retrieval.
Specifically, the developed MOON synchronously learns  the hash codes with multiple lengths in a unified framework.
Besides, to enhance the underlying discrimination, we combine the clues from the multimodal data, semantic label and learned hash codes for hash learning.
As far as we know, the proposed MOON is the first attempt to simultaneously learn different length hash codes without retraining in cross-media retrieval.
Experiments on several databases show that our MOON can achieve promising performance, outperforming some recent competitive shallow and deep methods.
\end{abstract}


\end{frontmatter}



\section{Introduction}
With the rapid development of smart devices and multimedia technologies, tremendous amount of data (\emph{e.g.,} texts, videos and images) are poured into the Internet every day \citep{chaudhuri2020cmir,cui2020efficient,zhang2020scalable,zhang2021dah,hu2019triplet,zhang2021fast}.
In the face of massive multimedia data, how to effectively retrieve the desired information with hybrid results (\emph{e.g.,} texts, images) becomes an urgent but intractable problem.
To this end, many research works have been devoted to cross-media retrieval.
The key challenge of cross-media similarity search is mitigating the “media gap”, because different modalities may lie in completely distinct feature spaces and have diverse statistical properties.
To erase the discrepancy, the basic solution is to project the multimedia data into a common subspace to learn the correspondence.
Besides, with the continuous increasing of multimedia data, the efficiency has become another obstacle restricting the application in real-world systems.
To solve this issues, binary coding (\emph{i.e.,} hashing technique), which encodes the heterogeneous data into a set of meaningful compact hash codes that keep the similarity and structure information in the original space, has been incorporated into the applications for cross-media similarity search.
Based on the binary encoding formulation, the retrieval can be efficiently performed with reduced storage cost.
With all these merits, therefore, hashing techniques have gained much attention, with many hashing based methods proposed for advanced cross-modal retrieval.

General speaking, existing cross-media hashing algorithms can be divided into two branches: unsupervised and supervised.
For the unsupervised models, representative approaches include, but not limited to, composite correlation quantization (CCQ) \citep{long2016composite}, collective matrix factorization hashing (CMFH) \citep{ding2014collective}, fusion similarity hashing (FSH) \citep{liu2017cross} and collective reconstructive embeddings (CRE) \citep{hu2018collective}.
Different from unsupervised approaches, supervised schemes can obtain more promising performance.
Typical supervised approaches include supervised matrix factorization hashing (SMFH) \citep{tang2016supervised}, which preserves the similarity by an $n\times n$ matrix and relaxes the discrete constraints to learn the binary representations.
Semantic preserving hashing (SePH) \citep{lin2015semantics} utilizes the KL-divergence and transforms the semantic information into probability distribution to learn the hash codes.
Because of the use of the $n\times n$ affinity matrix, SMFH and SePH are unscalable to large databases.
Semantic correlations maximization (SCM) \citep{zhang2014large} maximizes the correlations between different text modality and image modality to optimize the hash functions.
Label consistent matrix factorization hashing (LCMFH) \citep{wang2018label} proposes a novel matrix factorization framework and directly utilizes the supervised information to guide hash learning.
However, SMFH, SCM, SePH and LCMFH solve the binary constraints by a continuous scheme, leading to a large quantization error.
To this end, discrete cross-modal hashing (DCH) \citep{xu2017learning} directly embeds the supervised information into the shared subspace and learns the binary codes by a bitwise scheme.
Scalable matrix factorization hashing (SCARATCH) \citep{li2018scratch}, which learns a latent semantic subspace by adopting a matrix factorization scheme and generates hash codes discretely.
Discrete latent factor hashing (DLFH) \citep{jiang2019discrete}, which can effectively preserve the similarity information into the binary codes.
More recently, many deep hashing models have also been developed, such as adversarial cross-modal retrieval (ACMR) \citep{wang2017adversarial}, deep cross-modal hashing (DCMH) \citep{jiang2017deep} and self-supervised adversarial hashing (SSAH) \citep{li2018self}.
These methods usually obtain more promising performance compared with the shallow ones.

Although these algorithms have obtained satisfactory performance, there are still some limitations for advanced hashing models, which are introduced with our main motivations as below.
\begin{itemize}
\item In cross-modal retrieval, the existing methods usually predefine a fixed hash length (\emph{e.g.,} 8bits), such that we can utilize the 8bits query sample to retrieval the 8bits related ones.
However, when the hash length changes, the model needs to be retrained to learn the corresponding binary codes, which is inconvenient and cumbersome in real-world applications.
\item Most existing cross-modal approaches project the original multimedia data directly into hash space, implying that the binary codes can only be learned from the given original multimedia data.
However, the learned hash codes contain valuable semantic information, which has not been well explored.
Therefore, we propose to utilize the learned meaningful hash codes to assist in learning more discriminative binary codes.
\end{itemize}

To address the above issues, we develop a novel model for cross-media retrieval, \emph{i.e.,} multiple hash codes joint learning method (MOON).
In the proposed MOON, we can learn diverse length hash codes simultaneously, and the model does not need to be retrained when changing the length, which is very practical in real-world applications.
We can employ three types of information for hash learning (\emph{i.e.,} different length hash codes, semantic labels and the raw features of multimedia data).
In addition, we propose a bidirectional projection scheme to preserve more valuable information in the latent semantic subspace, with which the model can be efficiently optimized by developed optimization scheme and the binary codes can be generated discretely.
To the best of our knowledge, the proposed MOON is the first work to synchronously learn various length hash codes without retraining and is also the first attempt to utilize the learned hash codes for hash learning in cross-media retrieval.

The major contributions of this work are summarized as below.
\begin{itemize}
\item  We develop a novel framework, which can simultaneously learn different length hash codes without retraining. To our knowledge, this is the first work to explore multiple hash codes joint learning for cross-modal retrieval.
\item The learned various length hash codes are utilized for hash learning. The advantage is that the learned binary codes can be further explored to learn better binary codes.
\item We propose a bidirectional projection scheme to preserve more discriminative information, and the supervised semantic information can be fully utilized by the label reconstruction scheme. The optimization problem can be effectively addressed by the proposed optimization scheme without relaxation processing.
\end{itemize}

The rest of this article is organized as follows. The developed model is presented in Section \ref{section2}. The experimental results are reported in Section \ref{section3}. Finally, Section \ref{section4} concludes this work.

\section{The Proposed Method}
\label{section2}
\subsection{Notations}
We introduce the designed approach and perform the experiments on bimodal databases for simplicity, but the proposed model can be generalized in multimodal scenarios (more than two modalities).
Suppose $O_{train}=\{o_i\}_{i=1}^n$ is the training set with \emph{n} samples.
The \emph{i}-th instance is $o_i=\{x_i^{(1)},~x_i^{(2)}\}$.
$x_i^{(1)}$ and $x_i^{(2)}$ denote the text and image data points, respectively.
$X^{(t)}\in \mathbb{R}^{d_t\times n}$ stands for the feature matrix of text/image modality.
$d_t$ denotes the corresponding dimension.
$Y\in\{0,1\}^{c\times n}$ denotes label matrix, where \emph{c} means the number of categories.
$\|\cdot\|$ means the Frobenius-norm for matrix or $\ell_2$-norm for vector.
$sgn(\cdot)$ is the sign function.
In this work, we aim to learn \emph{K} different hash codes.
$B^k\in\{-1,1\}^{r_k\times n}$ denotes the \emph{k}-th binary code $(1\leq k \leq K)$, where $r_k$ is the length of the \emph{k}-th hash code.

\subsection{The Framework}
To reduce the semantic gap between different modalities, we first map the original feature of text ($X^{(1)}$) and image ($X^{(2)}$) into a latent shared semantic subspace. The objective is formulated as,
 \begin{equation}\label{eq1}
\sum_{k=1}^{K}\beta^k(\|U_{1f}^kX^{(1)}-S^k\|^2+\|U_{2f}^kX^{(2)}-S^k\|^2,
\end{equation}
where $S^k\in \mathbb{R}^{r_k\times n}$ is the latent common semantic representation of the \emph{k}-th item.
$U_{tf}^k\in\mathbb{R}^{d_t\times n}$ ($t=1,2$) denotes the projection matrix that projects $X^{(t)}$ to the latent semantic subspace.
Although the above formulation can learn underlying semantic representations, such operations (\emph{i.e.,} forward projection, $X^{(t)}\longrightarrow S^k$) may lose some valuable information.
To solve this issue, we propose a bidirectional projection strategy (\emph{i.e.,} $X^{(t)}\longrightarrow S^k$\&$S^k\longrightarrow X^{(t)}$ (back projection)). Then, the formulation is updated as,
 \begin{equation}\label{eq2}
 \begin{split}
 &\sum_{k=1}^{K}\beta^k(\|U_{1f}^kX^{(1)}-S^k\|^2+\|U_{2f}^kX^{(2)}-S^k\|^2) \\
&+\sum_{k=1}^{K}\alpha^k(\|U_{1b}^kS^k-X^{(1)}\|^2+\|U_{2b}^kS^k-X^{(2)}\|^2).
\end{split}
\end{equation}

The above formulation can preserve more discriminative information in the latent semantic representations.
To verify it, we take the \emph{k}-th semantic representation as an example.
Then, we have,
 \begin{equation}\label{eq3}
 \begin{split}
 &\beta^k(\|U_{1f}^kX^{(1)}-S^k\|^2+\|U_{2f}^kX^{(2)}-S^k\|^2)\\
&+\alpha^k(\|U_{1b}^kS^k-X^{(1)}\|^2+\|U_{2b}^kS^k-X^{(2)}\|^2).
\end{split}
\end{equation}

By introducing the reconstruction errors (\emph{i.e.,} $\xi_{fi}^{(t)}$, $\xi_{bi}^{(t)}$), we have,
 \begin{equation}\label{eq4}
 \begin{split}
 &x_i^{(t)}=U_{tb}^ks_i^k+\xi_{bi}^{(t)}, \\
&s_i^k=U_{tf}^kx_i^{(t)}+\xi_{fi}^{(t)}.
\end{split}
 \end{equation}

The $\ell_2$ distance between $s_i^k$ and $s_j^k$ can be written as
 \begin{equation}\label{eq6}
\|s_i^k-s_j^k\|=\|U_{tf}^k(x_i^{t} -x_j^{t})+(\xi_{fi}^{(t)}-\xi_{fj}^{(t)})\|.
 \end{equation}

Similarly, the $\ell_2$ distance between $x_i^{(t)}$ and $x_j^{(t)}$ can be written as
 \begin{equation}\label{eq7}
\|x_i^{(t)}-x_j^{(t)}\|=\|U_{tb}^k(s_i^k -s_j^k)+(\xi_{bi}^{(t)}-\xi_{bj}^{(t)})\|.
 \end{equation}

Based on the matrix triangle and cosine inequalities, we have,

 \begin{equation}\label{xxx1}
 \begin{split}
 \|x_i^{(t)}-x_j^{(t)}\| &\leq \|U_{tb}^k(s_i^k -s_j^k)\|+\|(\xi_{bi}^{(t)}-\xi_{bj}^{(t)})\|  \\
 &\leq\|U_{tb}^k\|~\|s_i^k -s_j^k\|+\|\xi_{bi}^{(t)}-\xi_{bj}^{(t)}\|.
 \end{split}
  \end{equation}
  \begin{equation}\label{xxx2}
  \begin{split}
 \|s_i^k-s_j^k\|& \leq \|U_{tf}^k(x_i^{t} -x_j^{t})\|+\|\xi_{fi}^{(t)}-\xi_{fj}^{(t)}\| \\
 &\leq \|U_{tf}^k\|~\|x_i^{t} -x_j^{t}\|+\|\xi_{fi}^{(t)}-\xi_{fj}^{(t)}\|.
 \end{split}
   \end{equation}

Furthermore, for ease of representation, we utilize $D_1$ and $D_2$ to represent $1/\|U_{tb}^k\|$ and $\|U_{tf}^k\|$, respectively, \emph{i.e.,} $D_1=1/\|U_{tb}^k\|$ and $D_2=\|U_{tf}^k\|$. Besides, $\psi_1(i,j)=\|\xi_{bi}^{(t)}-\xi_{bj}^{(t)}\|/\|U_{tb}^k\|$ and $\psi_2(i,j)=\|\xi_{fi}^{(t)}-\xi_{fj}^{(t)}\|$ are the bound errors. Then, we have,

 \begin{equation}\label{eq8}
 \begin{split}
&\|s_i^k-s_j^k\|\geq D_1\|x_i^{(t)}-x_j^{(t)} \|-\psi_1(i,j), \\
&\|s_i^k-s_j^k\|\leq D_2\|x_i^{(t)}-x_j^{(t)} \|+\psi_2(i,j).
\end{split}
 \end{equation}

Ideally, the reconstruction errors will vanish when $X^{(t)}$ and $S^k$ are perfectly factorized. Accordingly, $\psi_1(i,j)$ and $\psi_2(i,j)$ will be decreased to 0.
In effect, the bound errors (\emph{i.e.,} $\psi_1(i,j)$ and $\psi_2(i,j)$) are reduced in the process of optimization.
Then, $\|s_i^k-s_j^k\|$ is tight to the lower bound $D_1\|x_i^{(t)}-x_j^{(t)}\|$ and upper bound $D_2\|x_i^{(t)}-x_j^{(t)}\|$, that is, $D_1\|x_i^{(t)}-x_j^{(t)}\| \leq \|s_i^k-s_j^k\| \leq D_2\|x_i^{(t)}-x_j^{(t)}\|$, which is approximately bi-Lipschitz continuous. Thus, the similarity of original multimodal data can be well preserved for latent semantic representations by the bidirectional projection scheme.
According to the above analysis, we can prove that the bidirectional projection scheme can preserve richer information in the latent semantic representations compared with the one-side projection scheme.

After learning the latent semantic representations, we assume that the hash codes can be learned from the representations.
Then, the formulation is defined as below,
  \begin{equation}\label{eq10}
  \begin{split}
&\sum_{k=1}^{K}\|B^k-R^kS^k\|^2, \\
&s.t.~~ R^k{R^k}^T=I,~~B^k\in\{-1,1\}^{{r_k}\times n},
\end{split}
 \end{equation}
where $R^k\in\mathbb{R}^{r_k\times r_k}$ is the orthogonal matrix, keeping the discrete constraints in the process of training and directly learning the binary codes.

As discussed above, the learned hash codes contain rich semantic information. The learned hash codes with different lengths can be utilized for hash learning. To be specific, many works have demonstrated that longer hash codes can achieve better performance due to longer hash codes can be embedded more discriminative information \citep{shi2016kernel,wang2017survey}. This phenomenon indicates that the learned binary codes can be further re-utilized to learn better hash codes. Inspired by this, we further re-utilize the learned hash codes to learn powerful hash codes.
Then, we define the following formulation,
   \begin{equation}\label{eq11}
   \sum_{k=1}^{K-1}\|B^k-T^kB^{k+1}\|^2,
 \end{equation}
where $T^k \in \mathbb{R}^{r_k \times r_{k+1}}$ is a mapping matrix, $r_k$ and $r_{k+1}$ denote the length of $B^k$ and $B^{k+1}$, respectively.

In the above, the hash codes are learned from two kinds of information, \emph{i.e.,} the raw features of the multimodal data and the learned hash codes.
To further improve the search performance, the supervised semantic information is adopted in our model, assumed that the semantic label matrix can be reconstructed by the learned latent semantic representations.
Then, the loss is defined as.
    \begin{equation}\label{eq12}
   \sum_{k=1}^{K}\|Y-P^kS^{k}\|^2,
 \end{equation}
 where $P^k \in \mathbb{R}^{c \times r_k}$ is a mapping matrix, which is utilized to project latent semantic representations into semantic labels.

We combine Eqs.~(\ref{eq2}), (\ref{eq10}), (\ref{eq11}) and (\ref{eq12}) to formulate the following overall objective function,
    \begin{equation}\label{eq13}
\begin{split}
&\min_{B^k,U_{tf}^k,U_{tb}^k,S^k,R^k,T^k,P^k}\mathcal{L}(B^k,U_{tb}^k,U_{tf}^k,S^k,R^k,T^k,P^k) \\
&=\sum_{k=1}^{K}\beta^k(\sum_{t=1}^2\|U_{tf}^k\phi(X^{(t)})-S^k\|^2) \\
&+\sum_{k=1}^{K}\alpha^k(\sum_{t=1}^2\|U_{tb}^kS^k-\phi(X^{(t)})\|^2) \\
&+\sum_{k=1}^{K}\|B^k-R^kS^k\|^2+\sum_{k=1}^{K-1}\mu^k\|B^k-T^kB^{k+1}\|^2  \\
&+\sum_{k=1}^{K}\omega^k\|Y-P^kS^{k}\|^2+\lambda Reg(U_{tf}^k,U_{tb}^k,S^k,T^k,P^k), \\
&s.t.~~ R^k{R^k}^T=I,~~B^k\in\{-1,1\}^{{r_k}\times n},
\end{split}
 \end{equation}
where $Reg(\cdot)=\|\cdot\|^2$ is utilized for regularization.
$\beta^k$, $\gamma^k$, $\mu^k$ and $\lambda$ are the balance parameters. Besides, the RBF kernel is utilized to obtain the transformed representations $\phi(X^{(t)})$,  which has been widely utilized in many works \citep{kulis2009kernelized,shen2015supervised,li2018scratch}. Specifically, $\phi(X^{(t)})=[\phi(x_1^{(t)}),~\phi(x_2^{(t)}),\cdots,~\phi(x_n^{(t)})]$. The kernel features are represented as $\phi(x_i^{(t)})=[exp(-\|x_i^{(t)}-a_1^{(t)}\|^2/2\sigma_t^2),\cdots,exp(-\|x_i^{(t)}-a_m^{(t)}\|^2/2\sigma_t^2)]$. $\{a_j^{(t)}\}_{j=1}^m$ denotes the $m$ anchor points ($m<n$) that are randomly chosen from the training samples. The kernel width $\sigma_t=1/nm\sum_{i=1}^n\sum_{j=1}^{m}\|x_i^{(t)}-a_j^{(t)}\|$.

 \subsection{Optimization}
In this section, we develop an alternating optimization scheme to address the optimization issue in Eq.~(\ref{eq13}).
To clearly elaborate the optimization procedure, we only show the \emph{k}-th item.
The details are given as follows.

\textbf{Step-1:} To update $S^k$ with other variables fixed, let $\frac{\partial\mathcal{L}}{\partial S^k}=0$, the solution can be obtained as
    \begin{equation}\label{eq14}
    \begin{split}
&S^k=(\omega^k{P^k}^TP^k+\sum_{t=1}^2\alpha^k{U_{tb}^k}^TU_{tb}^k+{R^k}^TR^k+(2\beta^k+\lambda)I)^{-1} \\
&\bullet(\omega^k{P^k}^TY+{R^k}^TB^k+\sum_{t=1}^{2}(\alpha^k{U_{tb}^k}^T\phi(X^{(t)})+\beta^kU_{tf}\phi(X^{(t)}))),
\end{split}
 \end{equation}
 where $I$ is an identity matrix.

\textbf{Step-2:} To update $U_{tb}^k$ with other variables fixed, let $\frac{\partial\mathcal{L}}{\partial U_{tb}^k}=0$, we have
    \begin{equation}\label{eq15}
 U_{tb}^k=\alpha^k\phi(X^{(t)}){S^k}^T(\alpha_1^kS^k{S^k}^T+\lambda I).
 \end{equation}

\textbf{Step-3:} To update $U_{tf}^k$ with other variables fixed, let $\frac{\partial\mathcal{L}}{\partial U_{tf}^k}=0$, then obtain
    \begin{equation}\label{eq17}
 U_{tf}^k=\beta^kS^k{\phi(X^{(t)})}^T(\beta^k\phi(X^{(t)}){\phi(X^{(t)})}^T+\lambda I)^{-1}.
 \end{equation}

\textbf{Step-4:} To update $T^k$ with other variables fixed, let $\frac{\partial\mathcal{L}}{\partial T^k}=0$, then obtain
\begin{equation}\label{eq19}
T^k=\mu^kB^k{B^{k+1}}^T(\mu^kB^{k+1}{B^{k+1}}^T+\lambda I)^{-1}.
 \end{equation}

\textbf{Step-5:} To update $B^k$ with other variables fixed. Then, Eq.~(\ref{eq13}) can be simplified as
\begin{equation}\label{eq20}
\begin{split}
 &\|B^k-R^kS^k\|^2+\mu^k\|B^k-T^kB^{k+1}\|^2, \\
 &s.t.~~B^k\in\{-1,1\}^{r_k\times n}.
 \end{split}
 \end{equation}

The solution of $B^k$ can be easily obtained as
\begin{equation}\label{eq21}
B^k=sgn(R^kS^k+\mu^kT^kB^{k+1}).
 \end{equation}

 \textbf{Step-6:} To update $R^k$ with other variables fixed. We have,
 \begin{equation}\label{eq22}
 \begin{split}
\|B^k-R^kS^k\|^2,~~s.t.~{R^k}^TR^k=I.
\end{split}
 \end{equation}

Eq.~(\ref{eq22}) is a typical orthogonal procrustes issue \cite{schonemann1966generalized}, which can be easily addressed by SVD.
Then, we have $B^k{S^k}^T=W\Omega \overline{W}^T$, the solution can be obtained as
\begin{equation}\label{eq23}
R^k=W\overline{W}^T.
 \end{equation}

We summarize the optimization steps in Algorithm~\ref{alg1}.

\subsection{Out-of-Sample Extension}
In the query stage, the developed MOON can convert the multimodal data into hash codes.
Concretely, given a query data $x_{quey}^{(t)}$, the hash codes can be obtained by
\begin{equation}\label{eq24}
b_{query}^k=sgn(R^kU_{tf}^kx_{quey}^{(t)}),
 \end{equation}
where $R^k$ and $U_{tf}^k$ are learned in Algorithm~\ref{alg1}.

\begin{algorithm}[ht]
\caption{Muti-Hash Codes Jointly Learning (MOON)}
\label{alg1}
\textbf{Input}: Text/Image feature matrix $X^{(t)}~(t=1,2)$, label matrix $Y$, parameters $K$, $\{\alpha_t^k\}_{k=1}^K$, $\{\beta^k\}_{k=1}^K$, $\{\mu^k\}_{k=1}^K$, $\{\omega^k\}_{k=1}^K$ and $\lambda$.\\
\textbf{Output}: Rotation matrix $\{R^k\}_{k=1}^K$, hash matrix $\{B^k\}_{k=1}^K$ and $\{U_{tf}^k\}_{k=1}^K$.
\begin{algorithmic}[1] 
\STATE Calculate $\phi(X^{(t)})$;
\STATE Randomly initialize $B^1$, $R^1$, $S^1$, $U_{tf}^1$ and $P^1$; \\
\textbf{Repeat}
\STATE Update $\{S^k\}_{k=1}^K$ using Eq.~(\ref{eq14});
\STATE Update $\{U_{tb}^k\}_{k=1}^K$ using Eq.~(\ref{eq15}); \\
\STATE Update $\{U_{tf}^k\}_{k=1}^K$ using Eq.~(\ref{eq17}); \\
\STATE Update $\{T^k\}_{k=1}^K$ using Eq.~(\ref{eq19}); \\
\STATE Update $\{B^k\}_{k=1}^K$ using Eq.~(\ref{eq21}); \\
\STATE Update $\{R^k\}_{k=1}^K$ using Eq.~(\ref{eq23}); \\
\textbf{until} convergence or reach the maximum iteration. \\
\textbf{return}: $\{R^k\}_{k=1}^K$, $\{B^k\}_{k=1}^K$ and $\{U_{tf}^k\}_{k=1}^K$.
\end{algorithmic}
\end{algorithm}

\begin{table*}[htp]
\centering
\small
\caption{\label{tab2}mAP values of MOON and other approaches on three databases (The best results are marked in bold).}
\resizebox{\textwidth}{!}{
\begin{tabular}{llllllllllllll}
\hline
\multirow{2}{*}{Task}     & \multirow{2}{*}{Method} & \multicolumn{4}{c}{MIR Flickr}                                       & \multicolumn{4}{c}{IAPR-TC}                                          & \multicolumn{4}{c}{NUS-WIDE}                                         \\ \cline{3-14}
                          &                         & 12bits          & 24bits          & 36bits          & 48bits          & 12bits          & 24bits          & 36bits          & 48bits          & 12bits          & 24bits          & 36bits          & 48bits          \\ \hline \hline
\multirow{10}{*}{Img2Txt} & CMFH \citep{ding2014collective}                   & 0.5629          & 0.5628          & 0.5623          & 0.5624          & 0.3392          & 0.3399          & 0.3396          & 0.3398          & 0.3884          & 0.3937          & 0.3939          & 0.3900          \\
                          & SMFH \citep{tang2016supervised}                      & 0.6162          & 0.6180           & 0.6186          & 0.6203          & 0.3838          & 0.3893          & 0.3997          & 0.3986          & 0.4619          & 0.4664          & 0.4823          & 0.4873          \\
                          & SCM \citep{zhang2014large}                    & 0.6362          & 0.6389          & 0.6481          & 0.6517          & 0.3743          & 0.3881          & 0.3974          & 0.4010          & 0.5430          & 0.5525          & 0.5610           & 0.5626          \\
                          & GSPH \citep{mandal2017generalized}                    & 0.6593          & 0.6762          & 0.6793          & 0.6755          & 0.4123          & 0.4362          & 0.4445          & 0.4587          & 0.5788          & 0.5917          & 0.5901          & 0.5989          \\
                          & DCH \citep{xu2017learning}                     & 0.6702          & 0.6798          & 0.6795          & 0.6723          & 0.4457          & 0.4220           & 0.4299          & 0.4131          & 0.5751          & 0.5831          & 0.5619          & 0.5821          \\
                          & LCMFH \citep{wang2018label}                  & 0.6793          & 0.6848          & 0.6857          & 0.6878          & 0.4128          & 0.4207          & 0.4232          & 0.4257          & 0.6179          & 0.6313          & 0.6430          & 0.6475          \\
                          & CRE \citep{hu2018collective}                     & 0.6180          & 0.6160          & 0.6102          & 0.6084          & 0.4002          & 0.4179          & 0.4199          & 0.4229          & 0.4779          & 0.4877          & 0.4964          & 0.4969          \\
                          & SCRATCH \citep{li2018scratch}                 & 0.6985          & 0.6996          & 0.7064          & 0.7139          & 0.4472          & 0.4581          & 0.4676          & 0.4716          & 0.6109          & 0.6248          & 0.6304          & 0.6377          \\
                          & SLCH \citep{shen2020exploiting}                    & 0.6017          & 0.6190           & 0.6588          & 0.6308          & 0.3718          & 0.3797          & 0.4005          & 0.4161          & 0.5782          & 0.6004          & 0.5942          & 0.6263          \\
                          & LFMH \citep{donglin1}                    & 0.6809          & 0.6884           & 0.6919          & 0.6951          & 0.4231          & 0.4319          & 0.4396          & 0.4510          & 0.6130          & 0.6406          & 0.6492          & 0.6520          \\ \cline{2-14}
                          & \textbf{MOON (Ours)}           & \textbf{0.7139} & \textbf{0.7196} & \textbf{0.7195} & \textbf{0.7278} & \textbf{0.4653} & \textbf{0.4839} & \textbf{0.4938} & \textbf{0.4989} & \textbf{0.6299} & \textbf{0.6514} & \textbf{0.6584} & \textbf{0.6609} \\ \hline \hline
\multirow{10}{*}{Txt2Img} & CMFH \citep{ding2014collective}                     & 0.5553          & 0.5548          & 0.5543          & 0.5544          & 0.3427          & 0.3436          & 0.3432          & 0.3437          & 0.4135          & 0.4263          & 0.4241          & 0.4223          \\
                          & SMFH \citep{tang2016supervised}                    & 0.6319          & 0.6373          & 0.6388          & 0.6389          & 0.3949          & 0.4043          & 0.4192          & 0.4201          & 0.5343          & 0.5470           & 0.5696          & 0.5685          \\
                          & SCM \citep{zhang2014large}                      & 0.6218          & 0.6256          & 0.6321          & 0.6366          & 0.3694          & 0.3805          & 0.3878          & 0.3920          & 0.5125          & 0.5206          & 0.5311          & 0.5315          \\
                          & GSPH \citep{mandal2017generalized}                    & 0.6969          & 0.7172          & 0.7243          & 0.7299          & 0.4706          & 0.5069          & 0.5254          & 0.5441          & 0.6823          & 0.6989          & 0.7029          & 0.7091          \\
                          & DCH \citep{xu2017learning}                     & 0.7306          & 0.7422          & 0.7364          & 0.7260           & 0.5155          & 0.5147          & 0.5098          & 0.4993          & 0.6988          & 0.6978          & 0.6972          & 0.6977          \\
                          & LCMFH \citep{wang2018label}                   & 0.7336          & 0.7549          & 0.7508          & 0.7626          & 0.4383          & 0.4539          & 0.4665          & 0.4704          & 0.7391          & 0.7586          & 0.7657          & 0.7685          \\
                          & CRE \citep{hu2018collective}                     & 0.6125          & 0.6106          & 0.6049          & 0.6030           & 0.3961          & 0.4150          & 0.4178          & 0.4200          & 0.4685          & 0.4752          & 0.4856          & 0.4845          \\
                          & SCRATCH \citep{li2018scratch}                 & 0.7558          & 0.7569          & 0.7643          & 0.7783          & 0.5294          & 0.5638          & 0.5781          & 0.5919          & 0.7404          & 0.7607          & 0.7642          & 0.7695          \\
                          & SLCH \citep{shen2020exploiting}                   & 0.6281          & 0.6493          & 0.6924          & 0.6610          & 0.4006          & 0.4312          & 0.4501          & 0.4743          & 0.7076          & 0.7175          & 0.7135          & 0.7554          \\
                          & LFMH \citep{donglin1}                    & 0.7424          & 0.7502           & 0.7605          & 0.7676          & 0.4706          & 0.5044          & 0.5236          & 0.5362          & 0.7341          & 0.7584          & 0.7735          & 0.7738          \\ \cline{2-14}
                          & \textbf{MOON (Ours)}           & \textbf{0.7731} & \textbf{0.7793} & \textbf{0.7761} & \textbf{0.7822} & \textbf{0.5386} & \textbf{0.5749} & \textbf{0.5975} & \textbf{0.6074} & \textbf{0.7514} & \textbf{0.7722} & \textbf{0.7801} & \textbf{0.7839} \\ \hline
\end{tabular}}
\end{table*}

\subsection{Complexity analysis} \label{complexity}
In this subsection, we analyze the computational complexity of MOON.
To simplify the analysis, we only show the \emph{k}-th item, the results of other items are similar.
Concretely, the overall complexity includes $O(2fr_k^2+cr_k^2+2r_k^3+r_k(r_k+c+2f)n)$ for updating $S^k$, $O(fr_k^2+r_k^3+(fr_k+r_k^3)n)$ for updating $U_{tb}^k$, $O(cr_k^2+r_k^3+(cr_k+r_k^2)n)$ for solving $P^k$, $O(r_kr_{k+1}^2+r_{k+1}^3+(r_kr_{k+1}+r_{k+1}^2)n)$ for solving $T^k$, $O(r_kf^2+f^3+(fr_k+f^2)n)$ for updating $U_{tf}^k$, $O(r_k^3+r_k^2n)$ for $R^k$, $O((r_k+r_k^2)n)$ for $B^k$, respectively, where \emph{c} denotes the number of categories, $f$ is the anchor points and $r_k$ means the hash length.
$r_k,~r_{k+1}~c,~f\ll n$.
Thus, the overall complexity of MOON is linear to \emph{n} (the size of training data).
\begin{table}[ht]
\centering
\small
\caption{General statistics of three datasets.}
\resizebox{\linewidth}{!}{
\begin{tabular}{lccc}
\toprule
Statistics         & MIR Flickr  & IAPR-TC     & NUS-WIDE   \\ \midrule
Database size      & 20,015      & 20,000      & 186,577    \\
Training set size  & 18,015      & 18,000      & 10,000     \\
Query set size     & 2,000       & 2,000       & 2,000      \\
Retrieval set size & 18,015      & 18,000      & 184,577    \\
Label              & 24          & 255         & 10         \\
Image feature      & 512-D GIST  & 512-D GIST  & 500-D BOVW \\
Text feature       & 1,386-D BOW & 2,912-D BOW & 1000-D BOW \\ \bottomrule
\end{tabular}}
\label{tab1}
\end{table}
\section{Experiments}
\label{section3}
\subsection{Experimental settings}
\subsubsection{Datasets}
To testify the performance of our MOON, we perform extensive experiments on three large-scale databases, \emph{i.e.,} NUS-WIDE \citep{chua2009nus} IAPR-TC \citep{escalante2010segmented} and MIR Flickr \citep{huiskes2008mir}.
The statistics of these databases are reported in Table~\ref{tab1}.

\begin{figure*}[!h]
	\centering
	\includegraphics[width=6.9in]{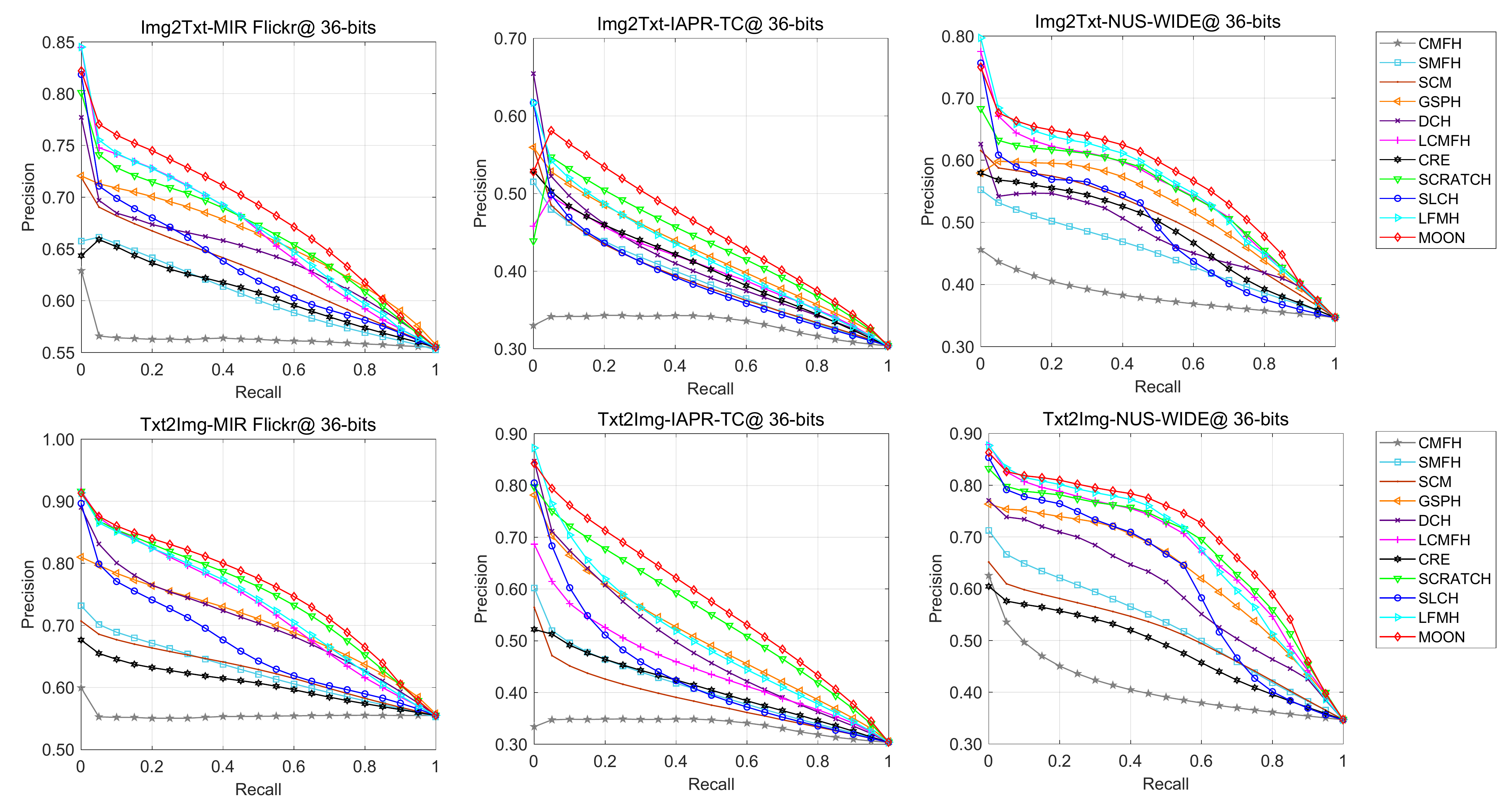}%
	\caption{PR-curves of different approaches on IAPR-TC, MIR Flickr and NUS-WIDE.}
	\label{Fig1}
\end{figure*}

\subsubsection{Baselines and Implementation details}
We compare the developed MOON with several recent competitive approaches.
For non-deep hashing methods, the compared approaches include SCM \citep{zhang2014large}, CMFH \citep{ding2014collective}, SMFH \citep{tang2016supervised}, DCH \citep{xu2017learning}, GSPH \citep{mandal2017generalized}, SCRATCH \citep{li2018scratch}, LCMFH \citep{wang2018label}, CRE \citep{hu2018collective}, SLCH \citep{shen2020exploiting} and LFMH \citep{donglin1}.
Specifically, CRE and CMFH are unsupervised models and the other algorithms are supervised.
For deep hashing algorithms, the baselines include DCMH \citep{jiang2017deep} and SSAH \citep{li2018self}, ADAH \citep{zhang2018attention}, PRDH \citep{yang2017pairwise}, AGAH \citep{gu2019adversary}, DADH \citep{bai2020deep} and MLCAH \citep{ma2020multi}.

In experiments, two typical cross-media retrieval tasks are conducted in this work, \emph{i.e.,} Img2Txt (using image to query related textual samples) and Txt2Img (using text to query related visual instances).
For the proposed MOON, we empirically set $\{\alpha^k\}_{k=1}^K=0.5$,  $\{\beta^k\}_{k=1}^K=1e3$,  $\{\mu^k\}_{k=1}^K=1e-6$, $\{\omega^k\}_{k=1}^K=1e3$ and $\lambda=5$. The number of anchor points (i.e., $m$) is set to 1000.
All experiments are carried out 5 times and the average value is taken as the final result.
In addition, two popular criteria, \emph{i.e.,} mean average precision (mAP) and precision-recall curve (PR-curve) are employed to evaluate the search performance.

\subsection{Comparison with Shallow Hashing methods}
\label{shallow}
In the experiments, four lengths of binary codes are adopted (\emph{i.e.,} \emph{K}=4) to testify the performance.
Specifically, the lengths are set to 12bits, 24bits, 36bits and 48bits, respectively.
The mAP results of our MOON and other compared algorithms are given in  Table~\ref{tab2}, in which the best results are shown in boldface.
From these results, it can be found that the developed MOON outperforms all the compared approaches in all cases, verifying the superiority of the developed MOON in cross-modal retrieval.
The superiority is mainly attributed to the joint learning of different hash codes in a unified framework, with three types of information, (\emph{i.e,} learned hash codes, semantic labels and raw features of multimodal data) being employed to obtain more discriminative hash codes.
In addition, the developed bidirectional projection scheme can preserve more effective information in the latent semantic subspace, where the optimization issue can be effectively addressed by the designed optimization scheme without relaxation.
For a detailed comparison, we also report the PR-curves on NUS-WIDE, MIR Flickr and IAPR-TC for Img2Txt and Txt2Img tasks, which are shown in Fig.~\ref{Fig1}.
As shown in the figure, the curves of our MOON are the highest in all cases, which indicates that the proposed MOON can obtain higher precision compared to the baselines at the same recall values.

\subsection{Comparison with Deep Hashing methods}
In this section, we set \emph{K}=3.
To be specific, the lengths are set to 16bits, 32bits and 64bits, respectively.
For fair comparison, the experimental settings of our model and other methods are the same as that of DCMH (please note that the experimental settings of this section are different from Section \ref{shallow}, more details can be referred to \citep{jiang2017deep}).
To be specific, we adopt CNN-F deep network pretrained on ImageNet to extract the image features, the text samples are described by the BOW features. For MIR Flickr database, 10,000 image-text pairs are arbitrarily
selected for training, and test set is built by randomly selecting 2,000 text-image pairs. For NUS-WIDE, 10,500 samples are selected for training and 2,100 data points are chosen for testing.
We conduct comparison with some recent deep methods on NUS-WIDE  and MIR Flickr.
Table~\ref{tab3} reports the experimental results.
It can be observed that the proposed MOON achieves good performance on NUS-WIDE and MIR Flickr, the performance is also superior to some comparative deep algorithms.
The above results further demonstrate the efficacy of the developed model.

\begin{table}[!ht]
\small
\caption{\label{tab3}mAP values of our method and other compared deep algorithms on NUS-WIDE and MIR Flickr.}
 \centering
  \resizebox{\linewidth}{!}{
\begin{tabular}{clcccccc}
\hline
\multirow{2}{*}{Task}    & \multirow{2}{*}{Method} & \multicolumn{3}{c}{NUS-WIDE}                       & \multicolumn{3}{c}{MIR Flickr}                     \\ \cline{3-8}
                         &                         & 16bits          & 32bits          & 64bits          & 16bits          & 32bits          & 64bits          \\ \hline \hline
\multirow{8}{*}{Img2Txt} & DCMH \citep{jiang2017deep}                  & 0.5912          & 0.6007          & 0.6090          & 0.7350          & 0.7371          & 0.7505          \\
                         & ADAH \citep{zhang2018attention}                   & 0.6698          & 0.6857          & 0.6993          & 0.7812          & 0.8033          & 0.8067          \\
                         & PRDH \citep{yang2017pairwise}                  & 0.4997          & 0.5373          & 0.4996          & 0.7422          & 0.7465          & 0.7560          \\
                         & SSAH \citep{li2018self}                    & 0.6420          & 0.6360          & 0.6390          & 0.7820          & 0.7901          & 0.8000          \\
                         & AGAH \citep{gu2019adversary}                   & 0.6455          & 0.6600          & 0.6502          & 0.7922          & 0.7946          & 0.8087          \\
                         & DADH \citep{bai2020deep}                   & 0.6492          & 0.6662          & 0.6665          & 0.8002          & 0.8070          & 0.8109          \\
                         & MLCAH \citep{ma2020multi}                   & 0.6334          & 0.6341          & 0.6347          & 0.7816          & 0.7918          & 0.8054          \\ \cline{2-8}
                         & MOON                     & \textbf{0.7698} & \textbf{0.7815} & \textbf{0.7884} & \textbf{0.8152} & \textbf{0.8221} & \textbf{0.8301} \\ \hline \hline
\multirow{8}{*}{Txt2Img} & DCMH \citep{jiang2017deep}                   & 0.6298          & 0.6411          & 0.6417          & 0.7637          & 0.7640          & 0.7752          \\
                         & ADAH \citep{zhang2018attention}                   & 0.6430          & 0.6249          & 0.6490          & 0.7463          & 0.7618          & 0.7637          \\
                         & PRDH \citep{yang2017pairwise}                  & 0.6616          & 0.6719          & 0.6677          & 0.7780          & 0.7694          & 0.7763          \\
                         & SSAH \citep{li2018self}                   & 0.6690          & 0.6620          & 0.6660          & 0.7910          & 0.7950          & 0.8030          \\
                         & AGAH \citep{gu2019adversary}                   & 0.6313          & 0.6425          & 0.6337          & 0.7886          & 0.7903          & 0.8045          \\
                         & DADH \citep{bai2020deep}                    & 0.6507          & 0.6678          & 0.6819          & 0.7920          & 0.7955          & 0.8046          \\
                         & MLCAH \citep{ma2020multi}                   & 0.6624          & 0.6732          & 0.6870          & 0.7841          & 0.7954          & 0.7957          \\ \cline{2-8}
                         & MOON                     & \textbf{0.7366} & \textbf{0.7457} & \textbf{0.7513} & \textbf{0.7942} & \textbf{0.7992} & \textbf{0.8119} \\ \hline
\end{tabular}}
\end{table}

\subsection{Further analysis }
\subsubsection{Ablation study}
 To confirm the contribution of different components of the proposed MOON, we further perform some experiments to evaluate the effectiveness of different components on NUS-WIDE. To be specific, we design several derivatives of our MOON: 1) MOON-L, which denotes that supervised information term is removed in MOON (\emph{i.e.,} $\omega^k$ is set to 0 ). 2) MOON-K, which means that we only utilize the raw features rather than the kernelized features. 3) MOON-H means that the learned hash codes are not utilized in MOON (\emph{i.e.,} $\mu^k$ is set to 0). This variant is utilized to testify the effect of using the learned hash codes. 4) MOON-B denotes the back projection terms are discarded. 5) MOON-S denotes the hash codes are learned separately by using the multimodal data and semantic label.
Table~\ref{tab4} reports the experimental results. We can see that the performance of MOON is much better than that of MOON-L, the main reason is that MOON-L cannot utilize the semantic information from labels, which can help to narrow the semantic gap between different modalities, further indicating the importance of supervised information. Besides, the performance of MOON is also superior to MOON-H, demonstrating that the learned longer hash codes can be reutilized to learn powerful hash codes. From the search results of MOON-B and MOON, we can see that MOON outperforms MOON-B, the reason is that the bidirectional projection strategy can preserve more valuable information compared with the one-side projection scheme. Moreover, MOON also outperforms MMON-K and MOON-S, which demonstrates the efficacy of kernelized features and joint learning.
According to the above analysis and the experimental results, we can see that our MOON outperforms MOON-H, MOON-L, MOON-K, MOON-S and MOON-A, respectively, demonstrating the importance of all the designed features of the developed model.

\begin{table}[!ht]
\small
  \caption{\label{tab4}mAP results of MOON and its variants on IAPR-TC.}
 \centering
  \resizebox{\linewidth}{!}{
\begin{tabular}{lcccccccl}
\hline
\multirow{2}{*}{Variants} & \multicolumn{4}{c}{Img2Txt}      & \multicolumn{4}{c}{Txt2Img}      \\ \cline{2-9}
                          & 12bits & 24bits & 36bits & 48bits & 12bits & 24bits & 36bits & 48bits \\ \hline \hline
MOON-L                    & 0.3696 & 0.3842 & 0.3946 & 0.3974 & 0.3775 & 0.3955 & 0.4085 & 0.4123 \\
MOON-K                    & 0.4427 & 0.4582 & 0.4662 & 0.4721 & 0.5234 & 0.5592 & 0.5780 & 0.5916 \\
MOON-S                    & 0.4245 & 0.4324 & 0.4416 & 0.4668 & 0.5110 & 0.5397 & 0.5657 & 0.5707 \\
MOON-H                    &0.4496  &0.4556  &0.4730  &0.4775  &0.5172  &0.5621  &0.5727  &0.5799  \\
MOON-B                    & 0.4560 & 0.4768 & 0.4820 & 0.4879 & 0.5325 & 0.5701 & 0.5823 & 0.5904 \\ \hline
MOON                      & \textbf{0.4653} & \textbf{0.4839} & \textbf{0.4938} & \textbf{0.4989} & \textbf{0.5386 }& \textbf{0.5749} & \textbf{0.5975} & 0.\textbf{6074 }\\ \hline
\end{tabular}}
\end{table}

\subsubsection{Time Cost Analysis}
In this section, some experiments are conducted on MIR Flickr to illustrate the efficiency of MOON. All experiments are tested 5 runs on a desktop computer with an 8-core\@3.2GHZ CPU and 32G RAM, the average results are recorded as the final values. Please note that our MOON learns different length hash codes simultaneously. For fair comparison, we should perform these compared approaches on different hash length settings (12bits+24bits+36bits), and the sum of the training time of different hash code lengths (12bits+24bits+36bits) is used as the final comparison time. Table \ref{tab5} shows the training time of all algorithms. From the results, it can be found that the developed MOON is much faster than these compared approaches. The main reason is that our MOON can simultaneously learn different length hash codes without retraining, while all compared methods need to be retrained to learn the corresponding hash codes. From the above analysis and results, we can conclude that the developed MOON can achieve good efficiency.

\begin{table}[!ht]
\small
 \centering
   \caption{\label{tab5}Training time on three datasets (seconds).}
   \resizebox{\linewidth}{!}{
\begin{tabular}{lccc}
\hline
Method/Dataset               & MIR Flickr(12+24+36) & IAPR-TC(12+24+36)  & NUS-WIDE(12+24+36) \\ \hline
CMFH                         & 12.15      & 13.84    & 84.18    \\
SMFH                         & 14789.34   & 21601.43 & 1686.94  \\
SCM                          & 29.84      & 44.30    & 20.00    \\
GSPH                         & 19746.36   & 20143.89 & 11003.07 \\
DCH                          & 649.14     & 2201.87  & 239.24   \\
LCMFH                        & 166.58     & 745.96   & 55.48    \\
CRE                          & 32.45      & 52.30    & 15.82    \\
SCRATCH                      & 29.77      & 89.05    & 11.24    \\
SLCH                         & 23.70      & 23.11    & 11.14    \\
LFMH                         & 160.30     & 710.22   & 49.75    \\ \hline
MOON                         & \textbf{11.32}      & \textbf{11.47}    & \textbf{6.22 }    \\ \hline
\end{tabular}}
\end{table}

\subsubsection{Visualization}
\begin{figure}[!h]
	\centering
	\includegraphics[width=3.6in]{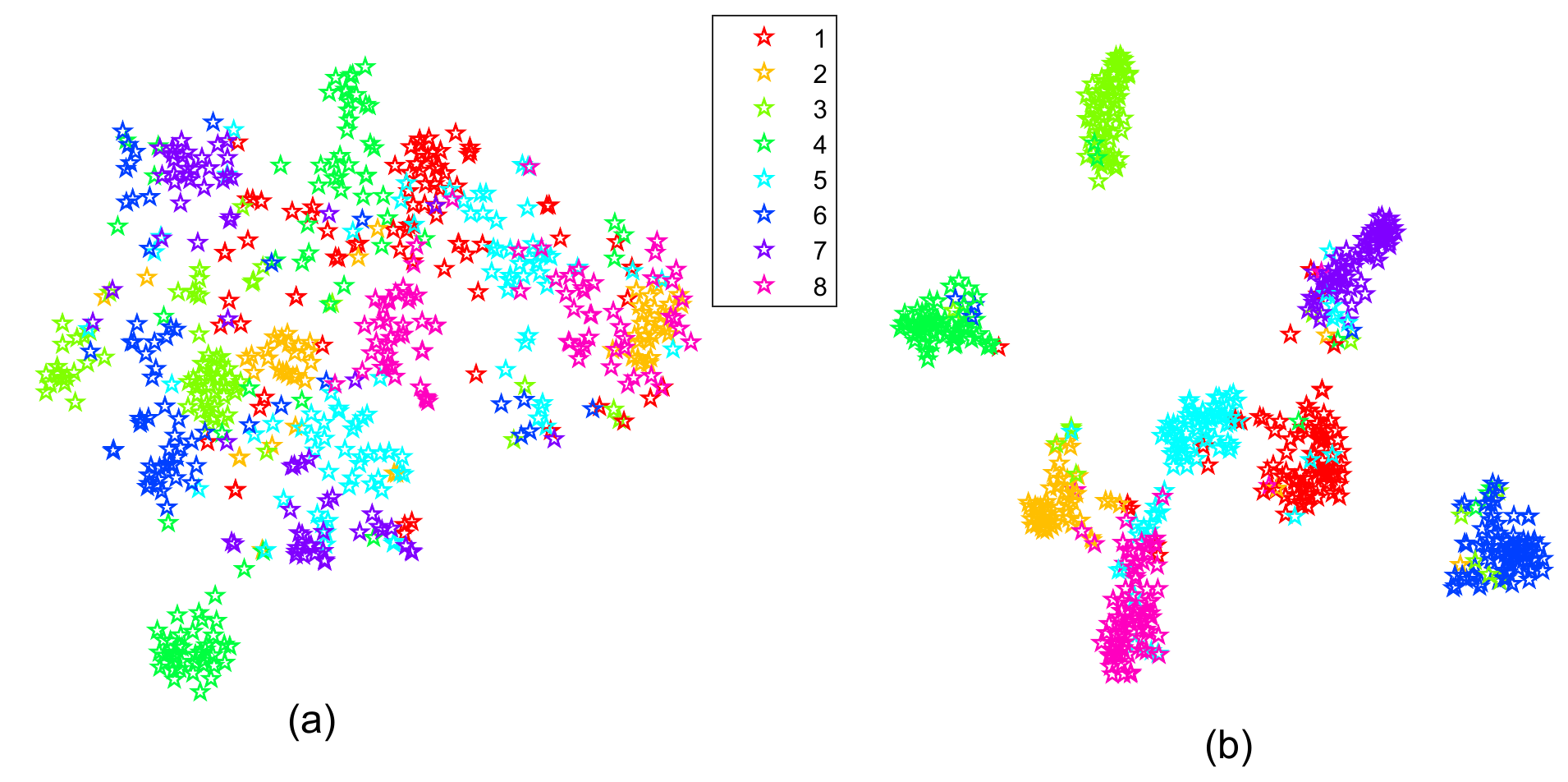}%
	\caption{(a) Raw features. (b) Learned semantic representations.}
	\label{Fig2}
\end{figure}

To further testify the efficacy of our model, we perform several visualization experiments on the popular LabelMe database \citep{russell2008labelme}.
Concretely, we arbitrarily select 300 textual samples and 300 visual ones, and choose the case when the code length is 36 to display.
For clear comparison, we first conduct PCA \citep{abdi2010principal} to get the same dimensional data.
The t-SNE tools \citep{maaten2008visualizing} are employed to visualize the distributions of raw text and image instances, which is shown in Fig.~\ref{Fig2} (a), it can be found the distribution is scattered and difficult to separate.
In addition, the learned semantic representations (without binarization) are plotted in Fig.~\ref{Fig2} (b).
From the figure, we can see that the learned semantic features of our MOON are formed by several different clusters, which further verifies the efficacy of the proposed MOON.

\section{Conclusion}
\label{section4}
In this paper, we develop a novel multiple hash codes joint learning method for cross-modal similarity search, called MOON.
Different from the existing methods, we can learn multiple hash codes with different lengths simultaneously in the proposed model.
Besides, the learned hash codes are utilized to learn more discriminative hash representations.
We also propose a bidirectional projection scheme to preserve more semantic information in the latent semantic subspace, and a label reconstruction strategy is adopted to capture the semantic structure.
Moreover, we further develop an alternative optimization scheme to effectively address the model without relaxation.
Experiments on several databases show that the performance of our method is superior against some recent shallow and deep models.
In addition, we find that with the increase of the number of hash codes, the number of parameters will also increase, which is a challenge for this model.
We will give an effective scheme to solve this problem in our future work.
\section*{Acknowledgments}
This work was supported by the NSFC [Grant 62020106012, U1836218, 61672265], UK EPSRC GRANT EP/N007743/1 and MURI/EPSRC/DSTL GRANT EP/R018456/1.
\small
\bibliographystyle{model2-names}
\bibliography{refs}

\end{document}